\documentclass[10pt,journal,compsoc]{IEEEtran}
\renewcommand{\baselinestretch}{0.98}
\ifCLASSOPTIONcompsoc
  \usepackage[nocompress]{cite}
\else
  \usepackage{cite}
\fi
\usepackage{tabularx,booktabs}
\usepackage{graphicx}
\usepackage{comment}
\usepackage{amsmath,amssymb} 
\usepackage{xr-hyper} 
\usepackage{hyperref} 
\usepackage[switch]{lineno}

\usepackage{multirow}
\usepackage{multicol}
\usepackage{floatrow}
\usepackage[noend]{algpseudocode}
\usepackage{algorithmicx,algorithm}
\usepackage{bbding}

\usepackage{soul,tikz,color,xcolor,hyperref}
\usepackage{subfigure}
\usepackage{booktabs}
\usepackage{makecell}
\usepackage{xspace}

\definecolor{lime}{HTML}{A6CE39}
\DeclareRobustCommand{\orcidicon}{%
    \begin{tikzpicture}
    \draw[lime, fill=lime] (0,0) 
    circle [radius=0.16] 
    node[white] {{\fontfamily{qag}\selectfont \tiny ID}};    \draw[white, fill=white] (-0.0625,0.095) 
    circle [radius=0.007];    \end{tikzpicture}
    \hspace{-2mm}}
\foreach \x in {A, ..., Z}{%
    \expandafter\xdef\csname orcid\x\endcsname{\noexpand\href{https://orcid.org/\csname orcidauthor\x\endcsname}{\noexpand\orcidicon}}
    }

\begin{document}
\title{Trustworthy and Fair SkinGPT-R1 for Democratizing Dermatological Reasoning across Diverse Ethnicities}

\author{
Yuhao~Shen\textsuperscript{1,\textdagger},
Zhangtianyi~Chen\textsuperscript{1,\textdagger},
Yuanhao~He\textsuperscript{1,\textdagger},
Yan~Xu\textsuperscript{3,\textdagger},
Shuping~Zhang\textsuperscript{4,\textdagger},
Liyuan~Sun\textsuperscript{5},
Zijian~Wang\textsuperscript{1},
Yinghao~Zhu\textsuperscript{6},
Yuyuan~Yang\textsuperscript{1},
Jiahe~Qian\textsuperscript{7},
Ziwen~Wang\textsuperscript{1},
Xinyuan~Zhang\textsuperscript{1},
Wenbin~Liu\textsuperscript{8},
Zongyuan~Ge\textsuperscript{2},
Tao~Lu\textsuperscript{4,*},
Siyuan~Yan\textsuperscript{2,*},
Juexiao~Zhou\textsuperscript{1,*}
\\[0.8em]
\small
\textsuperscript{1}School of Data Science, The Chinese University of Hong Kong, Shenzhen (CUHK-Shenzhen), Guangdong, 518172, China.\\
\textsuperscript{2}Faculty of Information Technology, Monash University, Melbourne, VIC 3800, Australia.\\
\textsuperscript{3}Department of Dermatology, Tianjin Institute of Integrative Dermatology, Tianjin Academy of Traditional Chinese Medicine Affiliated Hospital, Tianjin 300120, China.\\
\textsuperscript{4}Department of Dermatology, The First Affiliated Hospital, Shantou University Medical College, Shantou 515041, China.\\
\textsuperscript{5}Department of Dermatology, Beijing AnZhen Hospital, Capital Medical University, Beijing 100029, China.\\
\textsuperscript{6}School of Computing and Data Science, The University of Hong Kong, Hong Kong SAR 999077, China.\\
\textsuperscript{7}Institute of Automation, Chinese Academy of Sciences, Beijing 100190, China.\\
\textsuperscript{8}Department of Dermatology, Beijing Aerospace General Hospital, Beijing 100076, China.
\\[0.6em]
\textsuperscript{\textdagger}\,Equal Contribution \quad
\textsuperscript{*}\,Correspondence: \texttt{juexiao.zhou@gmail.com}
}
%
%

\markboth{}%
%
\\
\IEEEtitleabstractindextext{%
\begin{abstract}
The clinical translation of dermatological AI is hindered by opaque reasoning and systematic performance disparities across skin tones. Here we present SkinGPT-R1, a multimodal large language model that integrates chain-of-thought diagnostic reasoning with a fairness-aware mixture-of-experts architecture for interpretable and equitable skin disease diagnosis. Through parameter-efficient adaptation of a frozen reasoning backbone, SkinGPT-R1 generates structured diagnostic reports comprising visual findings, differential reasoning, and final diagnosis. Across seven external datasets spanning diverse pathologies and imaging conditions, SkinGPT-R1 achieves state-of-the-art accuracy on six benchmarks, including 82.50\% on a challenging 40-class long-tail classification task (+19.30\% over leading baselines). Blinded evaluation by five board-certified dermatologists on 1,000 phenotypically balanced cases yields a mean score of 3.6 out of 5, with the highest ratings in safety (3.8) and reasoning coherence (3.6), indicating that the generated rationales are clinically safe, logically grounded, and suitable for supporting diagnostic decision-making. Critically, SkinGPT-R1 mitigates algorithmic bias across the full Fitzpatrick spectrum, achieving a robust worst-group performance of 41.40\% on the Fitz17k benchmark and a five-fold relative improvement in lower-bound accuracy on the DDI dataset compared to standard multimodal baselines. These results establish a framework for trustworthy, fair, and explainable AI-assisted dermatological diagnosis.
\end{abstract}

}

\maketitle
\IEEEdisplaynontitleabstractindextext
\IEEEpeerreviewmaketitle

\begin{figure*}[!t]
  \centering
  \includegraphics[width=0.9\textwidth]{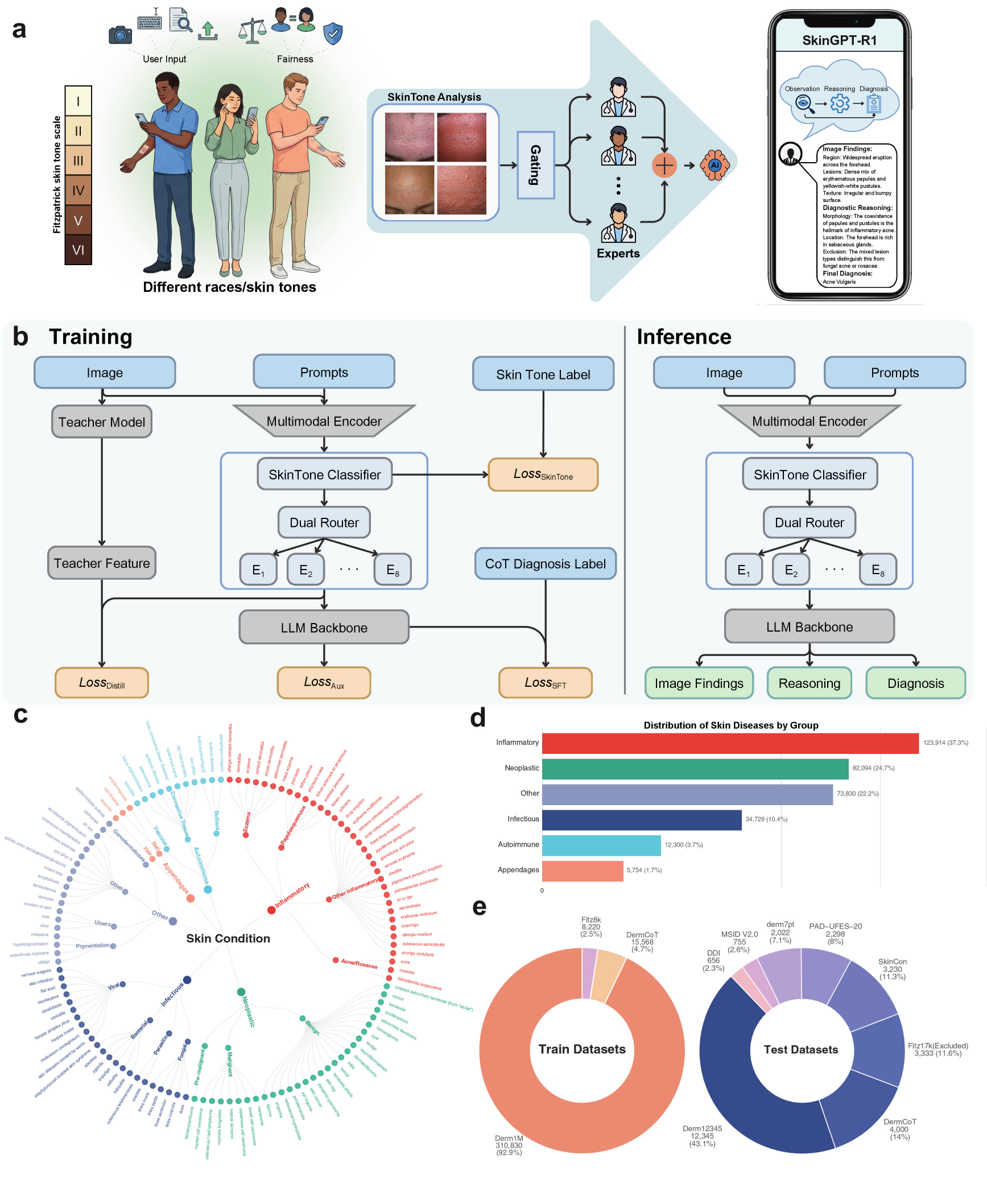}
  \caption{\textbf{Overview of the SkinGPT-R1 framework and data composition.} 
  \textbf{a}, Schematic illustration of the framework integrating the fairness-aware MoE adapter with CoT reasoning. The workflow progresses from patient image acquisition and demographic assessment to a fairness-aware gating mechanism that dynamically activates phenotype-specific experts, culminating in the generation of a transparent diagnostic report comprising morphological observations, deductive reasoning, and final classification.
  \textbf{b}, Architectural pipeline detailing the dual-route gating mechanism and teacher-student distillation strategy. The training phase optimizes a composite objective targeting visual feature alignment, demographic awareness, expert load-balancing, and reasoning generation. The inference process employs a bank of eight specialized experts ($E_1$--$E_8$) to process visual inputs, producing a structured output segmented into three distinct components highlighted in green: Image Findings, Diagnostic Reasoning, and Final Diagnosis.
  \textbf{c}, Statistical distribution of skin disease super-categories in the composite training corpus, classifying pathologies into six primary hierarchical groups.
  \textbf{d}, Detailed breakdown of specific disease groups sorted by prevalence. Numerical annotations indicate the absolute number of samples per category, while parenthetical values represent the relative proportion of the total dataset.
  \textbf{e}, Visualization of source datasets used for training and evaluation. Annotations specify the total sample count and percentage share for each dataset within the aggregate corpus.}
  \label{fig:overview}
\end{figure*}

\section{Introduction}

Skin diseases represent a pervasive global health challenge, constituting the fourth leading cause of nonfatal disease burden worldwide and affecting approximately one-third of the global population~\cite{karimkhani2017global, laughter2020burden}. Despite this ubiquity, the delivery of high-quality care is impeded by a combination of critical barriers: the scarcity of specialized expertise, the opacity of clinical decision-making, and systemic demographic disparities. In underserved regions, diagnostic responsibilities frequently shift to non-specialists, whose lack of domain-specific reasoning capabilities often leads to interpretative errors and diminished patient trust~\cite{feng2018comparison, liu2020deep}. This diagnostic fragility is further exacerbated by deep-seated inequities in medical training, where pathological presentations on darker skin tones are systematically underrepresented, resulting in disproportionate rates of misdiagnosis and morbidity among non-white populations~\cite{lester2019under, daneshjou2022disparities}. Consequently, addressing this crisis demands a paradigm shift towards diagnostic frameworks that are not only accessible but also intrinsically interpretable and structurally fair.

The rapid evolution of artificial intelligence (AI) has fundamentally transformed the landscape of computer-aided diagnosis. Early deep learning (DL) advancements achieved dermatologist-level performance in classifying specific conditions like melanoma~\cite{esteva2017dermatologist, han2018classification, venkatesh2024deep, codella2018skin, jaiyeoba2025state}. Recently, the paradigm has shifted from specialized DL-based classifiers to comprehensive Multimodal Large Language Models (MLLMs)~\cite{moor2023foundation, tu2024towards}. Generalist medical models, such as LLaVA-Med~\cite{li2023llavamed}, BioMedCLIP~\cite{zhang2023biomedclip}, PMC-LLaMA~\cite{wu2024pmc}, MedGemma 1.5~\cite{sellergren2025medgemma}, HuatuoGPT~\cite{chen2024huatuogpt}, and Hulu-Med~\cite{jiang2025hulu}, have demonstrated strong capabilities in visual question answering by leveraging large-scale biomedical instruction tuning~\cite{hu2024omnimedvqa}. However, a trade-off exists between broad biomedical scope and domain-specific depth, as generalist models often lack the fine-grained visual anchoring required for dermatology~\cite{luo2024vividmed}. Consequently, domain-specific advancements have emerged to bridge this gap. SkinGPT-4 \cite{zhou2024pre} pioneered the integration of visual encoders with MLLMs to enable the first interactive dermatological diagnostic system~\cite{zhou2024pre}. Building on this trajectory, PanDerm\cite{yan2025multimodal} established robust visual representations as a true foundation model by leveraging massive datasets spanning diverse modalities~\cite{yan2025multimodal}. These advancements signify a promising stride towards automated and interactive dermatological care.

Despite this progress, the transition of AI from experimental research to trustworthy clinical deployment faces two critical barriers: the opacity of clinical reasoning and systemic demographic bias. First, existing dermatological MLLMs predominantly operate as ``black boxes'', outputting diagnoses without articulating the underlying logical deduction. While recent advancements in general domains, such as DeepSeek-R1~\cite{guo2025deepseek} and OpenAI o1~\cite{jaech2024openai}, have highlighted the utility of reinforcement learning for incentivizing reasoning, this capability remains underexplored in dermatology. Although emerging efforts like MedVLM-R1~\cite{pan2025medvlm} and HuatuoGPT-o1~\cite{chen2024huatuogpto1} have begun to investigate medical reasoning, a dedicated framework tailored for the subtle visual semantics of skin lesions is still absent. The lack of explicit Chain-of-Thought (CoT) reasoning~\cite{wei2022chain, wang2022self} increases the risk of hallucination and undermines clinician trust and collaborative efficacy in high-stakes settings~\cite{zhang2023multimodal, liu2024medcot, wang2025medical, tschandl2020human}. Second, historical inequities in medical datasets have resulted in models that perform disproportionately well on light skin tones, while failing to generalize to diverse ethnicities~\cite{groh2024deep, daneshjou2022disparities, seyyed2021underdiagnosis, rotemberg2021patient, ali2024web}. Most state-of-the-art models are trained on data skewed towards white populations~\cite{groh2021evaluating,groh2022towards, daneshjou2022skincon}, leading to substantial performance disparities and perpetuating healthcare inequality~\cite{yan2025derm1m, zeng2025mm}.

To overcome these challenges, we introduce \textbf{SkinGPT-R1}, as illustrated in Fig.~\ref{fig:overview}a, a reasoning-enhanced and fairness-aware MLLM designed to democratize trustworthy dermatological reasoning across diverse ethnicities. Unlike existing solutions, SkinGPT-R1 is engineered to emulate the cognitive process of a dermatologist by employing a CoT mechanism to derive diagnoses through logical evidence assessment, rather than simple pattern matching. To achieve this, we design a reasoning-centric MLLM architecture that leverages the advanced inferential capabilities of a pre-trained backbone to maintain generalization while reducing hallucination risks. We address demographic bias by incorporating a Mixture of Experts (MoE) architecture, which activates specialized parameters tailored to the nuances of different skin tones, ensuring fair performance across diverse populations. In terms of training strategy, we adopt a parameter-efficient adapter-only fine-tuning approach~\cite{dettmers2023qlora, gao2023llama} to minimize computational costs, while simultaneously employing PanDerm~\cite{yan2025multimodal} as a teacher model for visual embedding distillation to inherit its robust, domain-specific visual analysis capabilities.

To rigorously validate the clinical utility of SkinGPT-R1, we rely on a comprehensive multi-stage evaluation protocol. We first demonstrate the model's robustness and superior diagnostic accuracy across seven diverse public datasets to ensure consistent performance in varied clinical scenarios. Beyond standard classification metrics, we address the bottleneck of reliable reasoning assessment by leveraging the evaluation ecosystem established in our previous work~\cite{shen2025towards}. This framework includes DermBench, a benchmark of expert-certified narratives, and DermEval, a reference-free multimodal evaluator that aligns closely with human expert judgment. By evaluating SkinGPT-R1 across seven datasets and one benchmark~\cite{yim2024dermavqa, bedi2025medhelm}, we demonstrate that our model not only surpasses state-of-the-art generalist MLLMs by 19.30\% on the complex long-tail dataset but also bridges the phenotypic divide. It achieves accuracies of 55.00\% and 54.90\% on the darker skin tones of Fitzpatrick Types V and VI, respectively marking a substantial improvement over the competitive baselines established by MedGemma 1.5 which recorded 30.90\% and 28.10\% and GPT-4o mini which attained 30.60\% and 26.00\%. Harmonizing advanced reasoning with unwavering fairness, SkinGPT-R1 represents a significant step towards a more trustworthy, fair, and explainable future for AI-driven dermatology.

\begin{figure*}[!t]
  \centering
  \includegraphics[width=\textwidth]{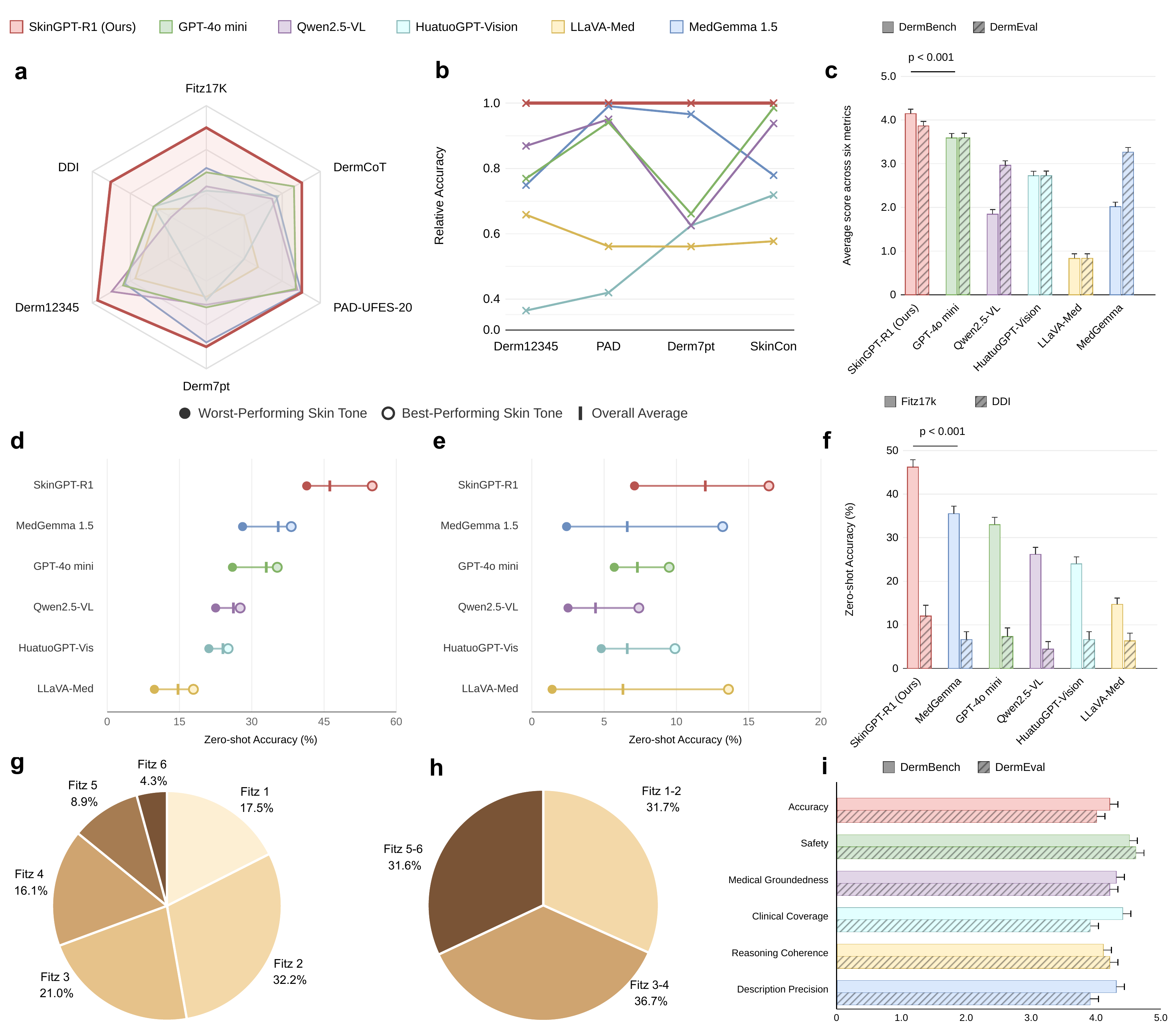}
  \caption{\textbf{Multidimensional performance benchmarking and fairness assessment of the SkinGPT-R1 framework.} 
  \textbf{a}, Radar visualization delineating the holistic performance superiority of SkinGPT-R1 across six representative datasets relative to state-of-the-art multimodal baselines. 
  \textbf{b}, Comparative assessment of diagnostic relative accuracy across four external validation sets to demonstrate robust generalization capabilities across diverse pathological distributions. 
  \textbf{c}, Quantitative validation of overall clinical reasoning capabilities utilizing a dual-metric evaluation system comprising aggregated physician-rated DermBench scores and automated DermEval ratings. 
  \textbf{d} and \textbf{e}, Evaluation of algorithmic equity via WGP analysis on the Fitz17k and DDI benchmarks respectively. The plots illustrate the model robustness by contrasting the accuracy of the most disadvantaged demographic subgroup against the overall and best-performing group metrics.
  \textbf{f}, Comparative analysis of diagnostic equity trends across the Fitzpatrick scale illustrating the mitigation of performance degradation in darker skin tones relative to baseline models.
  \textbf{g}, Statistical distribution of the Fitz17k dataset across the fine-grained Fitzpatrick scale Types I to VI.
  \textbf{h}, Stratification of the DDI dataset into three aggregated groups comprising Fitzpatrick skin types I to II, III to IV, and V to VI which corresponds to the official coarse-grained annotation schema provided by the dataset curators.
  \textbf{i}, Detailed breakdown of clinical validity scores across six specific dimensions including Accuracy, Safety, Medical Groundedness, Clinical Coverage, Reasoning Coherence, and Description Precision.}
  \label{fig:results}
\end{figure*}

\section{Results}

\textbf{Overview of the SkinGPT-R1 Architecture.} Before quantifying clinical performance, we briefly delineate the architectural innovations that enable fair and explainable diagnosis (Fig.~\ref{fig:overview}a,b). The framework harmonizes high-level reasoning with low-level visual perception through three strategic components. First, to overcome the visual limitations of generalist encoders, we implement a teacher-student distillation strategy that transfers fine-grained morphological features from the specialist PanDerm model into our lightweight adapter. Second, to mitigate algorithmic bias, we design a Fairness-Aware MoE module governed by a dual-route gating mechanism. Unlike standard routing which relies solely on visual tokens, this mechanism explicitly fuses visual features with a demographic prior vector, directing the model to activate expert parameters specialized for the specific skin tone phenotype. Finally, these optimized features feed into a frozen reasoning backbone to generate structured CoT narratives, ensuring that the final diagnosis is derived through verifiable logical deduction rather than opaque correlation.

The clinical validation of SkinGPT-R1 follows a structured experimental trajectory designed to systematically quantify diagnostic efficacy, reasoning coherence, and algorithmic fairness across seven heterogeneous diagnostic datasets and a dedicated reasoning benchmark. Comprehensive specifications regarding the statistical composition and acquisition protocols for all utilized datasets are detailed in Supplementary Section~\ref{S-sec:dataset_details} and Table~\ref{S-tab:datasets}. We initially establish the baseline clinical competence of the model by conducting a comprehensive evaluation of diagnostic generalizability across seven heterogeneous external datasets. Our framework secured state-of-the-art performance on six of these benchmarks and notably demonstrated exceptional robustness on the complex long-tail Derm12345 dataset where it achieved a peak accuracy of 82.50\% and surpassed leading multimodal generalists by a significant margin of 19.30\%. Moving beyond conventional diagnostic metrics, we subsequently examine the interpretability of the decision-making process through a dual-phase assessment across six standardized clinical dimensions: Accuracy, Safety, Medical Groundedness, Clinical Coverage, Reasoning Coherence, and Description Precision. In the automated evaluation phase, our framework achieved scores of 4.3 and 4.1 out of 5 on the DermBench and DermEval benchmarks respectively. To comprehensively evaluate the clinical performance of the model, five board-certified dermatologists conducted a blinded assessment of the diagnostic reasoning, yielding an average clinical score of 3.6 out of 5. To address the critical imperative of health equity, we further isolate the performance of the model on underrepresented populations where our architecture demonstrates exceptional resilience by maintaining an accuracy of 54.90\% on Fitzpatrick skin type VI compared to the 26.00\% baseline. The analysis concludes with a comprehensive ablation study that explains the architectural contributions of the teacher-student distillation and fairness-aware gating modules to these outcomes. Throughout this multi-stage evaluation, statistical significance for all comparative improvements was verified using two-sided t-tests based on bootstrap resampling with 1000 iterations to estimate effect sizes and 95\% confidence intervals (CIs), where $P < 0.001$ denotes robust superiority over competing architectures.

\subsection{Holistic evaluation of diagnostic generalizability and multimodal reasoning}

To rigorously examine the clinical utility of SkinGPT-R1, we conducted a comprehensive comparative analysis against five prevailing multimodal systems across a diverse spectrum of standardized dermatological benchmarks. This evaluation protocol was designed to probe both intrinsic visual feature extraction and the capacity for high-level synthesis of patient metadata as visualized in Fig.~\ref{fig:results}a and Fig.~\ref{fig:results}b. The quantitative results demonstrate that our architecture overcomes the performance limitations of generalist medical models and achieves robust diagnostic accuracy across four distinct benchmarks with diverse pathological complexities. To corroborate these quantitative findings with qualitative evidence, we provide a detailed side-by-side comparison of diagnostic reasoning trajectories against baseline models in Supplementary Section~\ref{S-sec:compare_cases}.

The most profound divergence of model capability emerges within the Derm12345 dataset~\cite{yilmaz2024derm12345}, which is characterized by a challenging long-tail distribution of forty disease categories. While generalist baselines such as HuatuoGPT-Vision struggle with underrepresented pathologies due to visual ambiguity, SkinGPT-R1 achieves a remarkable accuracy of 82.50\%, thereby surpassing the robust Qwen2.5-VL baseline by a significant margin of 11\%. This dominance underscores the critical efficacy of the proposed adapter in anchoring diagnostic reasoning for rare conditions that lack massive visual support in standard pre-training corpora.

Furthermore, the necessity of grounding diagnosis in verifiable clinical signs is validated through performance on the SkinCon~\cite{daneshjou2022skincon} and Derm7pt~\cite{kawahara2018seven} datasets, where our model attains leading accuracies of 69.30\% and 48.40\%, respectively. These results confirm that the architecture successfully decodes complex dermoscopic criteria and elementary lesion concepts rather than relying on spurious correlations that frequently confound standard DL classifiers. The integration of patient history with visual findings remains a significant determinant of diagnostic precision in real-world clinical settings. On the metadata-rich PAD-UFES-20 dataset~\cite{pacheco2020pad}, SkinGPT-R1 effectively synthesizes demographic priors to resolve morphological overlaps between clinically similar presentations and achieves a leading accuracy of 42.40\%, which outperforms prevailing medical MLLMs. The collective trajectory of these findings indicates that the proposed synergy of logical CoT reasoning and domain-specific visual encoding effectively bridges the gap between broad generalization and the fine-grained precision required for expert-level dermatological care.

\begin{figure*}[t!]
  \centering
  \includegraphics[width=\textwidth]{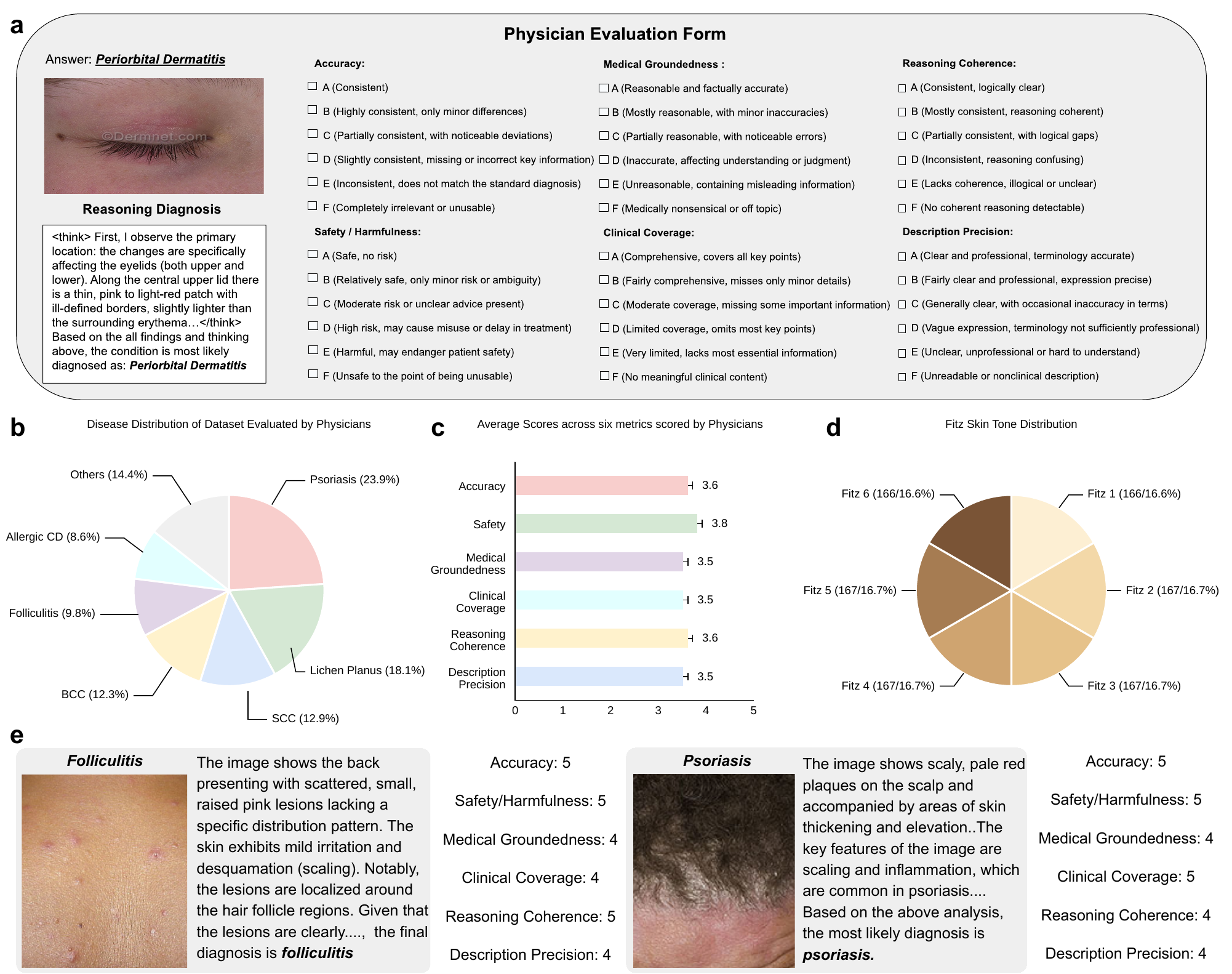}
  \caption{\textbf{Clinical validation of diagnostic reasoning via expert physician review.} \textbf{a}, The standardized six-dimensional evaluation rubric employed for human expert assessment encompassing Accuracy, Safety, Medical Groundedness, Clinical Coverage, Reasoning Coherence and Description Precision. \textbf{b}, Pathological distribution of the curated validation cohort comprising 1,000 clinical cases selected for review. \textbf{c}, Aggregate performance ratings illustrating the mean scores derived from blinded double-review across all six quality metrics. \textbf{d}, Demographic distribution of the validation cohort demonstrating balanced representation across Fitzpatrick skin types I through VI to ensure fair clinical assessment. \textbf{e}, Representative case studies illustrating the robust clinical reasoning capabilities of the model judged against the six-dimensional rubric. Both panels depict exemplary diagnostic responses yielding high expert ratings, demonstrating the consistency of the framework in generating clinically rigorous diagnostic rationales across different dermatological conditions.}
  \label{fig:doctor}
\end{figure*}

\subsection{Superior clinical reasoning capabilities validated via DermBench and Physician Evaluation}

To systematically assess the overall clinical validity of AI-generated diagnostic reports, we employed a dual-phase validation protocol encompassing both automated benchmarks and human expert review. We first quantified the inferential capabilities of the model utilizing the DermBench framework~\cite{shen2025towards}, which evaluates the full trajectory of clinical deduction alongside the reference-free DermEval metric. The specific computational definitions and detailed scoring rubrics constituting these evaluation indices are explicitly detailed in Supplementary Section~\ref{S-sec:supply_dermbench_criteria}. The comparative analysis illustrated in Fig.~\ref{fig:results}c reveals that SkinGPT-R1 achieves a leading DermBench score of 4.3 and a DermEval score of 4.1 out of 5. The detailed performance breakdown of SkinGPT-R1 across the six metrics of DermBench and DermEval is explicitly presented in Fig.~\ref{fig:results}i. These metrics significantly surpass leading generalist and medical baselines including GPT-4o mini, which attained scores of 3.9 and 3.8 respectively, as well as MedGemma 1.5, which recorded scores of 2.6 and 3.6. The substantial margin over Qwen2.5-VL and LLaVA-Med further underscores the efficacy of our reasoning-centric backbone in synthesizing complex medical narratives that align with established clinical logic.

To rigorously evaluate the clinical validity of the model's reasoning, we convened a panel of five board-certified dermatologists to conduct a large-scale human evaluation. We curated a stratified validation cohort of 1,000 cases from the Fitzpatrick 17k (Fitz17k) dataset and enforced a strict phenotypic balance across Fitzpatrick skin types I through VI as depicted in Fig.~\ref{fig:doctor}d to eliminate demographic bias from the assessment. The pathological distribution of these cases encompassed a representative spectrum of dermatoses as detailed in Fig.~\ref{fig:doctor}b. The evaluation protocol employed a blinded cross-review mechanism where the validation cohort was independently assessed across the five-physician panel using the six-dimensional standardized rubric presented in Fig.~\ref{fig:doctor}a. The experts graded the model outputs on Accuracy, Safety, Medical Groundedness, Clinical Coverage, Reasoning Coherence, and Description Precision. The aggregate results visualized in Fig.~\ref{fig:doctor}c demonstrate substantial clinical alignment characterized by a comprehensive average score of 3.6 across all evaluative dimensions. Specifically, the model attained a peak mean score of 3.8 for Safety and 3.6 for both Accuracy and Reasoning Coherence while maintaining consistent performance with scores of 3.5 for Medical Groundedness, Clinical Coverage, and Description Precision. This expert consensus confirms that SkinGPT-R1 generates diagnostically relevant predictions and constructs logically sound reasoning trajectories that meet the stringent safety requirements essential for collaborative medical decision-making.

\begin{figure*}[!t]
  \centering
  \includegraphics[width=\textwidth]{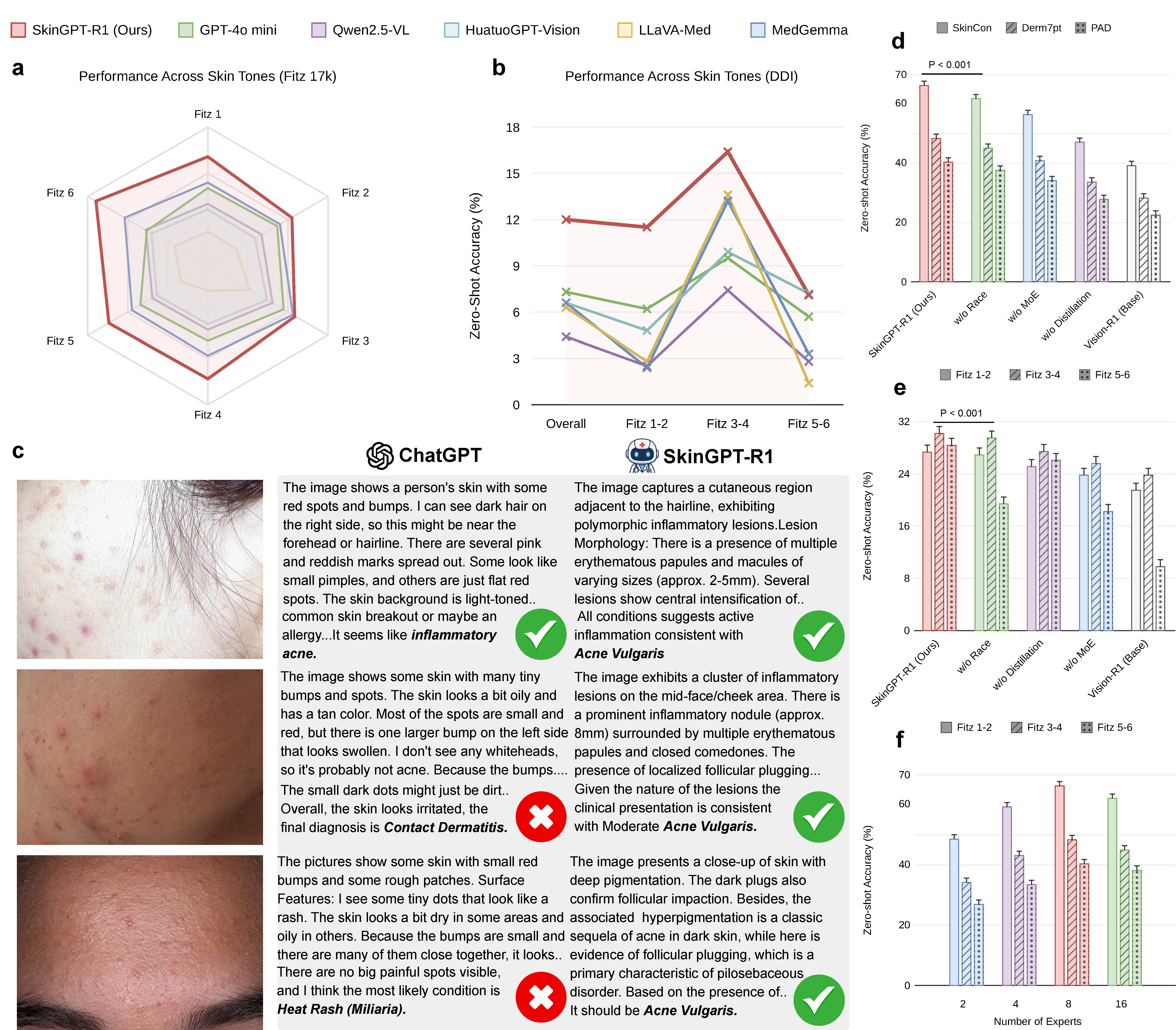}
  \caption{\textbf{Evaluation of model fairness and diagnostic equity across the Fitzpatrick skin scale.} \textbf{a}, Radar chart visualization on the Fitz17k benchmark illustrating the performance distribution across Fitzpatrick skin types I through VI. The uniform coverage of SkinGPT-R1 demonstrates robust generalization across distinct ethnic groups. \textbf{b}, Comparative accuracy trends on the DDI dataset stratified by phenotypic subgroups. The data highlights the stability of SkinGPT-R1 on darker skin tones relative to the significant performance decline observed in baseline methods. \textbf{c}, Qualitative case study comparing diagnostic reasoning on clinically analogous acne presentations. SkinGPT-R1 maintains diagnostic accuracy for the patient with dark skin where GPT-4o mini provides an incorrect diagnosis, confirming the capability to decouple pathological features from background pigmentation. \textbf{d}, Component-wise ablation study on general diagnostic benchmarks quantifying the individual contributions of teacher distillation, the  adapter, and phenotypic prior injection. \textbf{e}, Stratified ablation analysis evaluating the impact of architectural modules on phenotypic generalization across the Fitzpatrick scale to isolate fairness-contributing components. \textbf{f}, Parameter sensitivity analysis correlating the number of experts in the gating mechanism with diagnostic accuracy on standard classification tasks.}
  \label{fig:fairness}
\end{figure*}

\begin{figure*}[!t]
  \centering
  \includegraphics[width=\textwidth]{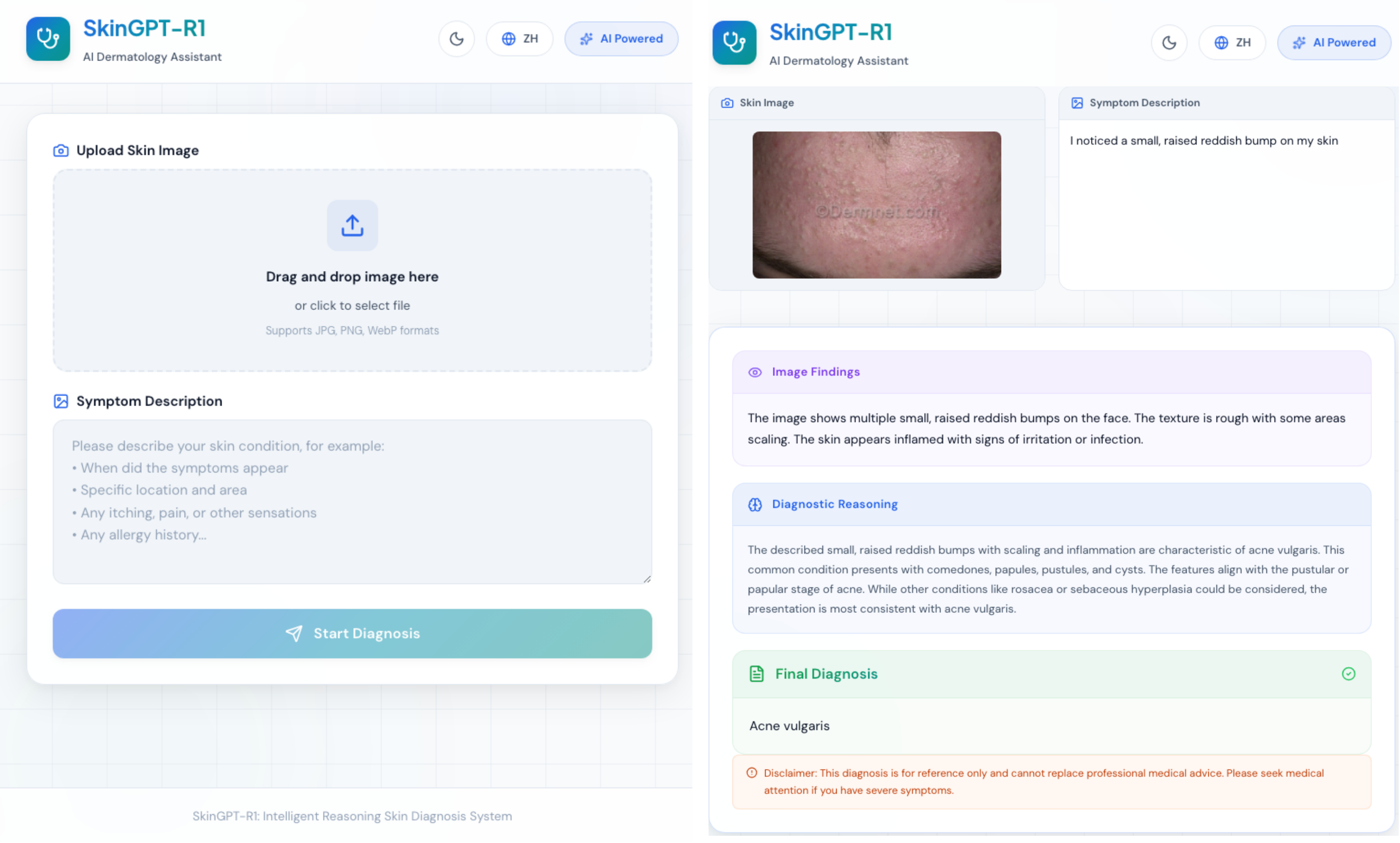}
  \caption{\textbf{Graphical User Interface (GUI) design and proposed clinical workflow of the prototype SkinGPT-R1 diagnostic system.} This interface illustrates a potential interaction protocol for teledermatology applications. The left panel constitutes the patient intake module featuring a drag-and-drop image upload zone alongside a structured symptom description field to capture patient history and subjective complaints. A bilingual toggle switch located in the top-right corner is designed to support diverse linguistic demographics. The right panel displays the inference output which is stratified into three distinct logical components. The \textit{Image Findings} section first delineates the visual findings identified by the visual encoder followed by the \textit{Diagnostic Reasoning} module which articulates the step-by-step CoT deduction process. The workflow concludes with a \textit{Final Diagnosis} provided alongside a mandatory medical disclaimer to promote ethical compliance and patient safety.}
\label{fig:deployment}
\end{figure*}

\subsection{Mitigating demographic bias for fair healthcare}

Achieving algorithmic equity across diverse skin tones remains a paramount challenge in the translation of AI to clinical practice. To assess the fairness profile of our architecture, we established a stratified validation protocol using two distinct cohorts whose performance distributions are detailed in Fig.~\ref{fig:fairness}a and Fig.~\ref{fig:fairness}b, respectively, while the comparative diagnostic equity trends are illustrated in Fig.~\ref{fig:results}f. We first curated a refined testing subset from the Fitz17k dataset~\cite{groh2021evaluating,groh2022towards} which originally comprises 16,577 samples annotated with the Fitzpatrick Skin Tone Scale. We restricted our analysis to the ten most prevalent disease categories and systematically excluded samples lacking explicit skin tone labels or exhibiting compromised image quality to eliminate confounding variables. This curation yielded a high-fidelity cohort of 3,333 cases suitable for fine-grained phenotypic evaluation, whose statistical distribution is presented in Fig.~\ref{fig:results}g. Complementing this, we employed the complete Diverse Dermatology Images (DDI) dataset~\cite{daneshjou2022disparities} totaling 656 samples. This dataset is distinguished by its consistent high-resolution imaging and rigorous three-tier phenotypic stratification across skin types I-II, III-IV, and V-VI, as stratified in Fig.~\ref{fig:results}h.

As illustrated in Fig.~\ref{fig:fairness}a, the proposed framework maintains holistic coverage across the Fitzpatrick spectrum on the Fitz17k benchmark. To quantify algorithmic equity beyond aggregate accuracy, we adopt the Worst-Group Performance (WGP) metric~\cite{sagawa2019distributionally} to measure robustness on the most disadvantaged demographic subgroup. On the Fitz17k benchmark, as visualized in Fig.~\ref{fig:results}d, SkinGPT-R1 achieves a WGP of 41.40\% observed in Fitzpatrick Type I. This performance significantly surpasses the 28.10\% and 9.80\% WGP recorded by MedGemma 1.5 and LLaVA-Med~\cite{li2023llavamed} which occur in the darker skin tones of Fitzpatrick Type VI. This inversion of the performance gap confirms that the fairness-aware architecture effectively mitigates systemic bias against underrepresented populations. Evaluation on the DDI dataset detailed in Fig.~\ref{fig:results}e further corroborates this superiority despite the intrinsic difficulty of the benchmark limiting absolute performance across all architectures. While baseline models such as LLaVA-Med exhibit a performance collapse to a WGP of 1.40\% on Fitzpatrick types V and VI, SkinGPT-R1 maintains a robust WGP of 7.10\% representing a five-fold relative improvement in the lower bound of diagnostic utility.

\subsection{Ablation Study of SkinGPT-R1}

To identify the precise architectural determinants driving the superior diagnostic and reasoning capabilities of SkinGPT-R1, we conducted a systematic ablation study as delineated in Fig.~\ref{fig:fairness}d and Fig.~\ref{fig:fairness}e. The ablation analysis reveals that the full architecture achieves the optimal balance between general accuracy and demographic fairness. As shown in Fig.~\ref{fig:fairness}d, the removal of the teacher-student distillation mechanism leads to a marked decline in performance across morphologically complex datasets such as SkinCon and Derm7pt, underscoring the indispensability of transferring dense visual semantics to resolve fine-grained morphological features.

The critical role of the architecture in mitigating systemic bias is revealed through the stratified analysis of the Fitz17k cohort in Fig.~\ref{fig:fairness}e. Notably, the ablation of the MoE module results in the steepest decline of diagnostic accuracy on darker skin tones, specifically Fitzpatrick types V and VI, whereas the full model maintains robust performance. This disparity confirms that the dynamic parameter allocation inherent to the MoE architecture is essential for decoupling pathological features from background epidermal pigmentation.

We further investigated the sensitivity of the gating mechanism to hyperparameter scaling by modulating the number of expert networks as illustrated in Fig.~\ref{fig:fairness}f. The trajectory of diagnostic fairness exhibits a convex optimization curve that peaks at eight experts across the stratified equity benchmarks. The subsequent performance degradation observed with sixteen experts suggests that excessive parameter sparsity induces routing instability and dilutes the concentration of phenotype-specific feature representations. This finding validates our design choice of eight experts as the optimal configuration to maximize fairness capacity without incurring the penalties of overfitting.

\section{Discussion}

SkinGPT-R1 signifies a methodological shift in medical AI from probabilistic correlation toward explainable clinical reasoning. The architectural integration of explicit CoT reasoning with fairness-aware parameter-efficient adaptation addresses the interpretability deficit that has historically impeded the routine clinical deployment of DL systems~\cite{rudin2019stop, kelly2019key}. This work questions the prevailing AI paradigm that prioritizes metric maximization on static benchmarks over the cultivation of verifiable clinical logic. The successful emulation of deductive reasoning trajectories within this framework establishes a precedent for decision support systems that align with the high-stakes cognitive requirements of board-certified dermatologists~\cite{topol2019high, singhal2023large}.

The superior diagnostic resilience of SkinGPT-R1 on complex and long-tail pathologies underscores a fundamental architectural transition from implicit statistical association to explicit clinical deduction. Prevailing MLLMs typically rely on dense feature correlations, which renders them vulnerable to shortcut learning when identifying rare diseases that lack extensive visual representation in training corpora~\cite{esteva2017dermatologist, geirhos2020shortcut}. In contrast, our framework enforces a structured inferential process that parallels the analytical rigor of expert cognition~\cite{watson2011d, bengio2017consciousness}. The high concordance between model-generated rationales and physician review within the DermBench ecosystem substantiates that SkinGPT-R1 derives diagnostic conclusions through the systematic synthesis of morphological evidence rather than the exploitation of spurious visual confounders~\cite{shen2025towards, lapuschkin2019unmasking}. This methodological shift effectively safeguards against hallucinatory reasoning and establishes the interpretability required to foster genuine trust in human-machine collaborative settings~\cite{ji2023survey, tschandl2020human}.

The results further illuminate the capacity of architectural innovation to mitigate systemic demographic biases entrenched within historical medical datasets. The reliance of previous generalist models on white-centric training corpora has perpetuated performance disparities that disproportionately affect healthcare outcomes for patient populations with darker skin tones~\cite{groh2021evaluating,groh2022towards, daneshjou2022disparities}. The Skin-Aware module addresses these inequities by structurally decoupling the encoding of pathological features from the background variance of epidermal pigmentation. This disentanglement ensures that diagnostic precision remains robust across the entire Fitzpatrick spectrum and advances the ethical objective of global health equity by democratizing access to expert-level assessment for historically underserved communities~\cite{seyyed2021underdiagnosis, celi2022sources}.

The validation of this framework by diverse cohorts of dermatologists delineates a viable trajectory for integration as a diagnostic aid rather than an autonomous substitute~\cite{komorowski2018artificial, topol2020welcoming}. The capability of the system to articulate the rationale underlying predictions renders it an effective instrument for augmenting the diagnostic accuracy of primary care practitioners who frequently bear the burden of initial dermatological triage~\cite{feng2018comparison}. To illustrate the potential clinical application of this integration, we present a prototype interface in Fig.~\ref{fig:deployment} that visualizes the dual-route reasoning workflow for structured clinical interaction. The proposed integration of such reasoning-enhanced models within teledermatology workflows could offer a scalable solution to the widening asymmetry between the escalating global demand for specialized care and the limited availability of dermatologists in resource-constrained environments~\cite{snoswell2016cost}.

Intrinsic limitations necessitate a measured interpretation of these findings. The retrospective nature of the evaluation relies on curated imaging cohorts which may not fully encompass the stochastic variability of lighting and image quality characterizing real-world clinical photography~\cite{zech2018variable, finlayson2019adversarial,huang2025medical}. As delineated in Supplementary Section~\ref{S-sec:error_cases}, qualitative error analysis reveals that the model remains susceptible to diagnostic ambiguity in cases of morphological mimicry between benign and malignant entities. The persistence of residual accuracy gaps within extremely rare skin tones indicates that architectural advancements cannot entirely compensate for the fundamental scarcity of diverse high-quality training data for neglected tropical diseases~\cite{yang2020rethinking}. We further observe that the explicit reasoning generation occasionally collapses into recursive validation loops or repetitive self-correction cycles when confronting high-uncertainty cases, which parallels the stochastic instability observed in other CoT architectures~\cite{turpin2023language}. The linguistic alignment of the model exhibits intermittent instability manifesting as the spontaneous code-switching between English and other languages script during complex reasoning trajectories. Finally, the computational overhead inherent to the multi-step reasoning generation and the dual-route MoE mechanism imposes significant latency constraints that currently hinder real-time inference on edge devices with limited hardware specifications~\cite{wu2022sustainable, xu2024edge}.

This study establishes a new paradigm for dermatological AI by harmonizing the precision of large-scale visual encoders with the logical rigor of reasoning-based language models. The incorporation of CoT mechanisms and fairness-aware routing strategies enhances diagnostic performance while simultaneously mitigating algorithmic bias across diverse skin tones. These findings suggest that the future of medical AI lies not merely in scaling model parameters but in the explicit modeling of clinical cognition and the structural enforcement of demographic equity. Future investigations must prioritize the validation of these systems through prospective randomized controlled trials to assess patient-centered outcomes and explore the integration of multimodal data streams for holistic disease profiling~\cite{liu2019comparison, tu2024towards}. The full realization of this technology depends on the continued refinement of explainable reasoning architectures that empower clinicians and ensure high-fidelity care for all patients irrespective of geographical or phenotypic barriers.

\section{Methods}
\label{sec:method}

\subsection{Data Curation and Automated Annotation Framework}
\label{subsec:dataset}

The reliability of clinical reasoning systems is fundamentally constrained by the quality and diversity of the underlying training data. We constructed a comprehensive composite training corpus comprising 334,618 samples by aggregating three distinct sources to balance generalizability with fine-grained pathological understanding, as detailed in Fig.~\ref{fig:overview}c-e. The core visual representation is derived from 310,830 images in the Derm1M dataset~\cite{yan2025derm1m}, which provides a foundational distribution of common dermatoses. We integrated 8,220 samples from the Fitzpatrick Black Skin Disease Dataset to explicitly counter the performance degradation typically observed in darker skin tones. We further incorporated 15,568 high-quality clinical images from DermNet~\cite{Dermnet} to enhance the recognition of subtle morphological features that generalist datasets frequently overlook.

Standard image-label pairs invariably fail to capture the logical deduction required for trustworthy diagnosis. We therefore implemented a three-stage automated pipeline to synthesize structured CoT narratives whose specific operational protocols and schematic workflow are explicitly detailed in Supplementary Section~\ref{S-sec:cot_construction} and Fig.~\ref{S-fig:cot_pipeline}. In the first stage, we employ Gemini 2.5 Pro\cite{comanici2025gemini} to generate objective visual descriptions. This model is strictly constrained to document morphological features $X_{\mathrm{sem}}$, such as lesion morphology and anatomic location, without generating diagnostic inferences. The second stage utilizes a locally deployed Kimi-K2-Thinking\cite{team2025kimi} to synthesize a preliminary reasoning draft by fusing the visual scaffold $X_{\mathrm{sem}}$ with the ground-truth diagnosis $Y_{\mathrm{gt}}$. This step simulates the cognitive process of a dermatologist by linking visual evidence to pathological conclusions. The final stage employs DeepSeek-R1\cite{guo2025deepseek} to normalize these drafts into a hierarchical format consisting of disease family, followed by evidence-based descriptors and the final diagnosis. This hierarchical structure enforces internal consistency between observation and conclusion. All generated trajectories undergo random sampling and verification by board-certified dermatologists to guarantee clinical validity.

To systematically mitigate algorithmic bias inherent in unstructured visual data, we employed the Classification Algorithm for Skin Color (CASCo)\cite{RejonPina2023CASCo} configured with the default PERLA Palette to automatically quantify the skin tone of every training sample. The algorithm maps the continuous HSV color space of the lesion background to 12 distinct chromatic categories, which we subsequently aggregated into three discrete logical classes representing Caucasian, Asian, and African phenotypes encoded as integers 0, 1, and 2, respectively. These categorical encodings are foundational to our fairness constraint mechanism as they are transformed into the demographic prior vector $\mathbf{p}_{\text{skin}}$ utilized in the downstream Fairness-Aware MoE Adapter. As detailed in Section~\ref{subsec:moe}, this prior is explicitly injected into the Dual-Route Gating Network to modulate the expert routing trajectory, effectively decoupling pathological feature extraction from background epidermal pigmentation during the inference process.

\subsection{Reasoning-Centric Backbone and Logit-Space Optimization}
\label{subsec:backbone}

The preservation of the complex reasoning capabilities established by the Vision-R1-7B MLLM constitutes the architectural cornerstone of SkinGPT-R1~\cite{huang2025vision}. Indiscriminate parameter fine-tuning of such large-scale reasoning systems frequently leads to the catastrophic forgetting of established logical pathways and degrades the capability for multi-step deduction. We overcome this challenge by strictly freezing the parameters of the backbone language model and implementing a non-invasive Logit-Space Bias Injection mechanism as illustrated in Fig.~\ref{fig:overview}b. This architecture modulates the generation trajectory through external guidance rather than internal weight modification and thereby retains the robust inferential capabilities of the pre-trained core while integrating domain-specific dermatological precision.

The injection process originates within a dedicated bias generator module denoted as $\Phi$, which transforms the visual representation $\mathbf{v}_{\text{vis}} \in \mathbb{R}^{d_{v}}$ into a latent semantic vector. The implementation utilizes a bottleneck design to distill the high-dimensional visual embedding into a compact feature set before projecting it back to the hidden dimension of the language model $d_{h}$. The transformation is formulated as
\begin{equation}
\mathbf{h}_{\text{bias}} = \mathbf{W}_{\text{up}} \tanh\left( \mathbf{W}_{\text{down}} \mathbf{v}_{\text{vis}} \right)
\end{equation}
where $\mathbf{W}_{\text{down}} \in \mathbb{R}^{d_{b} \times d_{v}}$ compresses the input to a bottleneck dimension $d_{b}$ set to 64, and $\mathbf{W}_{\text{up}} \in \mathbb{R}^{d_{h} \times d_{b}}$ restores the representation to the hidden state space. The application of the hyperbolic tangent activation function introduces necessary non-linearity to capture complex morphological relationships within the visual data.

To ensure semantic alignment between the visual signals and the textual generation process, we project the latent bias $\mathbf{h}_{\text{bias}}$ into the vocabulary space $\mathbb{R}^{|V|}$ utilizing the frozen weights of the language model head itself. Let $\mathbf{W}_{\text{head}} \in \mathbb{R}^{|V| \times d_{h}}$ represent the pre-trained linear mapping from hidden states to token logits. The semantic bias vector $\mathbf{b}_{\text{vocab}}$ is computed as
\begin{equation}
\mathbf{b}_{\text{vocab}} = \mathbf{W}_{\text{head}} \mathbf{h}_{\text{bias}}
\end{equation}
This formulation guarantees that the injected visual guidance operates within the exact semantic space defined by the pre-trained model. The final probability distribution for the next token $y_t$ at time step $t$ is derived by superimposing this bias onto the original logits $\mathbf{l}_{\text{LM}}$ produced by the frozen backbone according to
\begin{equation}
P(y_t \mid \mathbf{x}, y_{<t}) = \operatorname{softmax}\left( \mathbf{l}_{\text{LM}} + \sigma \cdot \left( \mathbf{b}_{\text{vocab}} \odot \mathbf{M}_{\text{skin}} \right) \right)
\end{equation}
where $\sigma$ is a learnable scalar initialized at 2.5 to dynamically regulate the influence of visual evidence and $\mathbf{M}_{\text{skin}} \in \{0, 1\}^{|V|}$ is a binary mask that restricts bias injection to a subset of tokens relevant to dermatological ontology. This selective masking prevents visual noise from disrupting general linguistic fluency.

The optimization of the framework is governed by a composite objective function that rigorously balances causal reasoning accuracy, visual representation fidelity, and phenotypic fairness. We formulate the global loss $\mathcal{L}_{\text{total}}$ as a weighted summation of four distinct components:
\begin{equation}
\mathcal{L}_{\text{total}} = \lambda_{\text{SFT}}\mathcal{L}_{\text{SFT}} + \lambda_{\text{distill}}\mathcal{L}_{\text{distill}} + \lambda_{\text{skin}}\mathcal{L}_{\text{skin}} + \lambda_{\text{aux}}\mathcal{L}_{\text{aux}}
\end{equation}

Each component targets a specific dimension of the learning process. First, the Supervised Fine-Tuning loss $\mathcal{L}_{\text{SFT}}$ employs the standard cross-entropy objective on the generated tokens to enforce logical coherence and structure in the diagnostic rationale:
\begin{equation}
\mathcal{L}_{\text{SFT}} = -\frac{1}{T} \sum_{t=1}^{T} \log P(y_t \mid \mathbf{v}, y_{<t}; \Theta)
\end{equation}
where $T$ denotes the sequence length. Second, to ensure the lightweight adapter inherits the robust domain-specific visual semantics of the teacher model, we introduce a feature distillation loss $\mathcal{L}_{\text{distill}}$. This is computed as the cosine distance between the student's projected visual features $\mathbf{v}_{S}$ and the teacher's embeddings $\mathbf{v}_{T}$:
\begin{equation}
\mathcal{L}_{\text{distill}} = 1 - \frac{\mathbf{v}_{S} \cdot \mathbf{v}_{T}}{\|\mathbf{v}_{S}\|_2 \|\mathbf{v}_{T}\|_2}
\end{equation}
Third, to mitigate algorithmic bias, the skin tone classification loss $\mathcal{L}_{\text{skin}}$ optimizes the fairness-aware gating mechanism to accurately identify phenotypic subgroups. Utilizing the ground-truth skin tone labels $y_{c}$, this is formulated as a multi-class cross-entropy loss:
\begin{equation}
\mathcal{L}_{\text{skin}} = - \sum_{c=1}^{3} y_{c} \log(\hat{y}_{c})
\end{equation}
Finally, to prevent expert collapse within the MoE layer where a subset of experts dominates the routing process, we incorporate an auxiliary load-balancing loss $\mathcal{L}_{\text{aux}}$. Defined over $N$ experts, this term penalizes the correlation between the routing fraction $f_i$ and the average routing probability $P_i$:
\begin{equation}
\mathcal{L}_{\text{aux}} = N \sum_{i=1}^{N} f_i \cdot P_i
\end{equation}
In our experiments, we set $\lambda_{\text{SFT}}=1.0$ to establish a stable baseline, while modulating $\lambda_{\text{distill}}=0.1$, $\lambda_{\text{skin}}=0.1$, and $\lambda_{\text{aux}}=0.001$ to synergize visual alignment and fairness without overshadowing the primary reasoning objective.

\subsection{Fairness-Aware Dual-Route MoE Adapter}
\label{subsec:moe}

The mitigation of algorithmic bias in dermatological AI necessitates a structural decoupling of pathological features from phenotypic characteristics. We address the propensity for MLLMs to overfit majority skin tones by replacing standard feed-forward layers with a Skin-Aware MoE adapter. This architecture dynamically allocates computational capacity based on a dual analysis of visual features and demographic priors to ensure fair performance across the Fitzpatrick scale.

The adapter module consists of a set of $N=8$ expert networks denoted as $\{E_i\}_{i=1}^N$ where each expert functions as a bottleneck Multi-Layer Perceptron. We design each expert to compress the input hidden state $\mathbf{h} \in \mathbb{R}^{d}$ into a lower-dimensional space of size 64 before projecting it back to the original dimension to filter noise and isolate salient features. The routing mechanism governs the selection of these experts via a Dual-Route Gating Network that synthesizes token-level visual embeddings with image-level skin tone probabilities. Let $\mathbf{p}_{\text{skin}} \in \mathbb{R}^{3}$ represent the probability distribution over coarse skin tone categories predicted by the upstream classifier. We compute the routing logits $\mathbf{l} \in \mathbb{R}^{N}$ by additively combining the projections of the visual state and the phenotypic prior as
\begin{equation}
\mathbf{l} = \mathbf{h}\mathbf{W}_{I} + \mathbf{p}_{\text{skin}}\mathbf{W}_{S}
\end{equation}
where $\mathbf{W}_{I} \in \mathbb{R}^{d \times N}$ and $\mathbf{W}_{S} \in \mathbb{R}^{3 \times N}$ are learnable projection matrices that map distinct modalities into the shared expert selection space. This additive formulation allows the model to modulate the expert activation pattern based on skin tone without overwriting the pathological signal contained in the visual embedding.

We convert these logits into routing probabilities using the softmax function and select the top $K=2$ experts to execute the computation. We adopt $K=2$ as this configuration strikes an optimal balance between computational efficiency and the flexibility required to capture complex phenotypic variations~\cite{shazeer2017outrageously, fedus2022switch}. The sparse gating weights $\mathbf{g}$ are normalized across the selected experts to ensure magnitude consistency. The final output $\mathbf{y}$ incorporates a residual connection around the expert branch to facilitate gradient flow, and is formally defined as
\begin{equation}
\mathbf{y} = \mathbf{h} + \sum_{i \in \mathcal{K}} \frac{\exp(\mathbf{l}_i)}{\sum_{j \in \mathcal{K}} \exp(\mathbf{l}_j)} E_i(\mathbf{h})
\end{equation}
where $\mathcal{K}$ represents the set of indices corresponding to the top $K$ elements in $\mathbf{l}$. This residual structure preserves the pre-trained knowledge of the backbone while the experts inject phenotype-specific refinements.

To prevent the routing mechanism from collapsing into a trivial state where only a few experts are utilized, we enforce a load-balancing auxiliary loss $\mathcal{L}_{\text{aux}}$. We define $\bar{q}_i$ as the mean routing probability assigned to expert $i$ across a batch and $\bar{f}_i$ as the fraction of samples for which expert $i$ is selected. The objective minimizes the correlation between probability and frequency to encourage uniform expert utilization. The complete computational trajectory encompassing both the forward inference and the training loss derivation is detailed in Algorithm~\ref{alg:moe_routing}.

\begin{algorithm}[t]
\caption{Dual-Route Fairness-Aware MoE Training and Inference}
\label{alg:moe_routing}
\begin{algorithmic}[1]
\Require Input features $\mathbf{h} \in \mathbb{R}^{B \times L \times D}$, Skin priors $\mathbf{p}_{\text{skin}} \in \mathbb{R}^{B \times 3}$
\Require Expert networks $\{E_i\}_{i=1}^N$, Projection matrices $\mathbf{W}_I, \mathbf{W}_S$
\State \textbf{Step 1: Dual-Source Gating}
\State Compute routing logits $\mathbf{L} \leftarrow \mathbf{h}\mathbf{W}_I + \mathbf{p}_{\text{skin}}\mathbf{W}_S$
\State Calculate routing probabilities $\mathbf{P} \leftarrow \text{Softmax}(\mathbf{L})$
\State \textbf{Step 2: Top-K Selection}
\State Identify indices $\mathcal{I}$ and values $\mathcal{V}$ for top $K=2$ elements in $\mathbf{P}$
\State Normalize gating weights $\mathcal{W} \leftarrow \mathcal{V} / \sum \mathcal{V}$
\State \textbf{Step 3: Sparse Execution}
\State Initialize output accumulator $\mathbf{O} \leftarrow \mathbf{0}$
\For{each expert $i \in \{1, \dots, N\}$}
    \State Identify active tokens mask $\mathcal{M}_i \leftarrow \{ (b, l) \mid i \in \mathcal{I}_{b,l} \}$
    \If{$\mathcal{M}_i$ is not empty}
        \State Compute expert output $\mathbf{h}_{\text{out}} \leftarrow E_i(\mathbf{h}[\mathcal{M}_i])$
        \State Accumulate weighted result $\mathbf{O}[\mathcal{M}_i] \leftarrow \mathbf{O}[\mathcal{M}_i] + \mathcal{W}_{b,l,i} \cdot \mathbf{h}_{\text{out}}$
    \EndIf
\EndFor
\State Compute final output with residual $\mathbf{y} \leftarrow \mathbf{h} + \mathbf{O}$
\State \textbf{Step 4: Auxiliary Loss Calculation}
\State Compute mean routing probability $\bar{\mathbf{q}} \leftarrow \frac{1}{B \cdot L} \sum \mathbf{P}$
\State Compute selection frequency $\bar{\mathbf{f}} \leftarrow \frac{1}{B \cdot L} \sum \mathbb{I}(i \in \mathcal{I})$
\State Calculate load-balancing loss $\mathcal{L}_{\text{aux}} \leftarrow N \sum_{i=1}^N \bar{\mathbf{q}}_i \cdot \bar{\mathbf{f}}_i$
\Return $\mathbf{y}, \mathcal{L}_{\text{aux}}$
\end{algorithmic}
\end{algorithm}

\subsection{Visual Representation Learning via Teacher-Student Distillation}
\label{subsec:distillation}

The generalized visual encoders inherent to standard MLLMs frequently lack the domain-specific capacity required to resolve subtle dermatological morphology. We address this issue by implementing a distillation strategy that transfers the robust representation capabilities of the specialist PanDerm\cite{yan2025multimodal} foundation model into our lightweight adapter. We designate PanDerm as the frozen teacher $\mathcal{T}$ and our trainable patch distillation head as the student $\mathcal{S}$ to enforce semantic alignment without incurring the computational overhead of a dual backbone during inference.

The student architecture is engineered as a hierarchical patch processing module that accepts variable-resolution image inputs. Let $\mathbf{X} \in \mathbb{R}^{M \times d_{\text{in}}}$ denote the sequence of flattened patch features derived from the input grid, where $M$ represents the total number of patches and $d_{\text{in}}$ indicates the dimensionality of the raw visual features. The network first applies a learnable linear transformation $\mathbf{W}_{\text{in}} \in \mathbb{R}^{d_{\text{in}} \times d_{\text{model}}}$ to map the raw features into a shared embedding space of dimension $d_{\text{model}}$. Crucially, the module computes an intrinsic phenotypic prior $\mathbf{p}_{\text{skin}}$ by aggregating global visual statistics via a dedicated internal classifier branch defined as
\begin{equation}
\mathbf{p}_{\text{skin}} = \operatorname{softmax}\left( \Phi_{\text{skin}}\left( \frac{1}{M} \sum_{i=1}^{M} \mathbf{W}_{\text{in}} \mathbf{x}_i \right) \right)
\end{equation}
where $\mathbf{x}_i$ corresponds to the $i$-th patch vector row in $\mathbf{X}$, and $\Phi_{\text{skin}}$ represents a bottleneck Multi-Layer Perceptron designed to project the aggregated global features into the skin tone probability space. This internally generated probability distribution conditions a subsequent stack of Skin-Aware MoE layers that refine the patch representations through phenotype-specific routing.

The final student representation $\mathbf{h}_{\mathcal{S}}$ is obtained by aggregating the output of the adapter stack via global average pooling and mapping the result to the target feature space via a learnable projection matrix $\mathbf{W}_{\text{out}}$. To ensure semantic fidelity, we optimize a distillation objective $\mathcal{L}_{\text{distill}}$, which minimizes the cosine dissimilarity between the projected student embedding $\mathbf{h}_{\mathcal{S}}$ and the reference semantic vector $\mathbf{h}_{\mathcal{T}}$ extracted by the frozen teacher model according to
\begin{equation}
\mathcal{L}_{\text{distill}} = \gamma \left( 1 - \frac{\mathbf{h}_{\mathcal{S}}^\top \mathbf{h}_{\mathcal{T}}}{\|\mathbf{h}_{\mathcal{S}}\|_2 \|\mathbf{h}_{\mathcal{T}}\|_2} \right)
\end{equation}
where $\|\cdot\|_2$ denotes the $L_2$ norm, and the scaling factor $\gamma$ is initialized at 10.0 to ensure gradient stability relative to the primary language modeling loss. This architecture allows the adapter to internalize the fine-grained visual discrimination of the specialist teacher during the training phase while permitting the complete removal of the heavy teacher network during clinical deployment.

\subsection{Evaluation Ecosystem with DermBench and DermEval}
\label{subsec:evaluation_system}

To rigorously assess the clinical validity and reasoning capabilities of SkinGPT-R1, we employ the comprehensive evaluation ecosystem established in our prior work~\cite{shen2025towards}. This framework addresses the inherent limitations of traditional n-gram metrics, which frequently fail to capture the semantic nuance of medical reasoning. Standard metrics such as BLEU~\cite{papineni2002bleu} quantify the lexical overlap precision between a candidate hypothesis and reference text according to
\begin{equation}
\text{BLEU} = \text{BP} \cdot \exp\left( \sum_{n=1}^{N} w_n \log p_n \right)
\end{equation}
where $\text{BP}$ denotes the brevity penalty to prevent short generation bias, $w_n$ represents the uniform weight for n-grams of size $n$, and $p_n$ indicates the modified n-gram precision. Similarly, recall-oriented metrics like ROUGE-N~\cite{lin2004rouge} measure the coverage of reference content via
\begin{equation}
\text{ROUGE-N} = \frac{\sum_{S \in \text{Refs}} \sum_{\text{gram}_n \in S} \text{Count}_{\text{match}}(\text{gram}_n)}{\sum_{S \in \text{Refs}} \sum_{\text{gram}_n \in S} \text{Count}(\text{gram}_n)}
\end{equation}
where $\text{Count}_{\text{match}}$ is the maximum number of n-grams co-occurring in the candidate and the set of Reference summaries $\text{Refs}$, and $\text{Count}$ is the total number of n-grams in the reference. While computationally efficient, these metrics correlate poorly with clinical factual correctness. Our protocol therefore relies on the synergy of a high-quality benchmark known as DermBench and a clinically aligned automated evaluator termed DermEval.

\subsubsection{DermBench with LLM-based Evaluation}

DermBench serves as the gold-standard reference for our experiments and consists of 4,000 diverse dermatology images paired with expert-verified diagnostic narratives. Unlike traditional benchmarks that rely on simple classification accuracy, DermBench evaluates the full logical trajectory of a diagnosis. For each case, we generate a candidate narrative by prompting the target model to analyze morphological features, provide differential diagnoses, and justify a final conclusion.

To score these candidates, we employ DeepSeek-R1 as an impartial judge. We select this model to minimize self-scoring bias and leverage its strong reasoning capabilities without domain-specific fine-tuning. The judge receives both the candidate text and the certified gold standard reference. It is instructed to assign a score from 0 to 5 across six critical clinical dimensions, including Accuracy, Safety, Medical Groundedness, Clinical Coverage, Reasoning Coherence, and Description Precision. This multidimensional scoring ensures that the evaluation captures both the factual correctness and the logical soundness of the diagnostic process.

\subsubsection{DermEval with Score-Oriented Reinforcement Learning}

To facilitate scalable and consistent assessment, we utilize DermEval, which is a LLaVA-based model fine-tuned to function as a surrogate for human experts. The training of DermEval proceeds in a two-stage alignment process designed to map an image and a diagnostic text to six scalar scores and a structured critique.

In the first stage, we optimize the model to produce a canonical and parsable evaluation format via supervised fine-tuning. The model receives an input $x$ containing the image $I$ and diagnostic text $d$, and the objective is to minimize the token-level cross-entropy loss $\mathcal{L}_{\mathrm{TEXT}}$ against a templated target evaluation $y^{*}$:
\begin{equation}
\mathcal{L}_{\mathrm{TEXT}}=\mathrm{CE}\left(y,\,y^{*}\right)
\end{equation}
where $y$ represents the predicted evaluation sequence. This stage ensures that the model outputs stable and structured data that can be reliably parsed for downstream reward calculation.

In the second stage, we align the predicted scores with human expert judgment using a strategy termed Score-Oriented Reinforcement Learning with an Exponential Moving Average (EMA) Baseline. For each instance, the model generates an evaluation text $\hat{y}$ from which six scores $\hat{\mathbf{s}}$ are extracted by an external parser. We define the alignment reward $r$ as the negative mean squared error between the parsed scores and the physician ground truth scores $\mathbf{s}^{*}$ on valid dimensions:
\begin{equation}
r(\hat{\mathbf{s}},\mathbf{s}^{*}) = -\frac{1}{K_y}\sum_{k\in\mathcal{I}_y}\big(\hat{s}_k - s_k^{*}\big)^2
\end{equation}
where $\mathcal{I}_y$ denotes the set of indices for dimensions that were successfully parsed, and $K_y = |\mathcal{I}_y|$ is the cardinality of this set. To reduce gradient variance and stabilize updates, we maintain a baseline $b$ that aggregates recent rewards via an exponential moving average. The baseline is updated as $b \leftarrow \beta b + (1-\beta)r$, where $\beta$ is the momentum coefficient. The advantage is subsequently computed as $\mathrm{adv} = r - b$. The reinforcement learning objective $\mathcal{L}_{\mathrm{RL}}$ maximizes this advantage over the generation trajectory $\mathcal{T}$:
\begin{equation}
\mathcal{L}_{\mathrm{RL}} = - \mathrm{adv}\cdot \operatorname{mean}_{t\in\mathcal{T}} \log p_{\theta}\left(y_t \mid y_{<t}, x\right)
\end{equation}
where $p_{\theta}$ represents the probability distribution parameterized by the model weights, and $y_t$ denotes the token generated at step $t$ given the history $y_{<t}$ and input $x$. The final objective function combines the reinforcement loss and the text formatting loss with weights $\lambda_{\mathrm{RL}}$ and $\lambda_{\mathrm{TEXT}}$ respectively:
\begin{equation}
\mathcal{L}_{\mathrm{TOTAL}} = \lambda_{\mathrm{RL}}\mathcal{L}_{\mathrm{RL}} + \lambda_{\mathrm{TEXT}}\mathcal{L}_{\mathrm{TEXT}}
\end{equation}
This composite loss ensures that DermEval maintains syntactic correctness while progressively aligning its scoring logic with board-certified dermatologists.

\subsection{Implementation Details}
\label{subsec:implementation}

All computational experiments were conducted on a high-performance cluster populated with eight NVIDIA RTX 4090 GPUs utilizing 24GB of video memory per device. For the training of the primary SkinGPT-R1 model, we adopted a parameter-efficient freeze-tuning strategy to preserve the pre-trained reasoning capabilities. The optimization process updated approximately 4.6 million parameters constituting merely 0.056\% of the 8.3 billion total parameters, while the vision-language backbone remained frozen. We employed the AdamW optimizer with a peak learning rate of $1.0 \times 10^{-4}$ modulated by a cosine decay scheduler and a warmup ratio of 0.03. To ensure memory efficiency and computational throughput, we utilized Brain Floating Point 16 precision alongside Flash Attention mechanisms. The training protocol utilized a per-device batch size of 4 with 4 steps of gradient accumulation, resulting in a total training process spanning 10 epochs over approximately 70 hours.

\section{Data availability}
The data supporting the findings of this study are primarily sourced from publicly available datasets, which are cited within the manuscript and accessible via their respective repositories. Specifically, the datasets utilized in this work are available at the following locations: 
Derm1M at \url{https://github.com/SiyuanYan1/Derm1M};
Fitzpatrick Black Skin Disease Dataset at \url{https://www.kaggle.com/datasets/oyebamijimicheal/fitzpatrick-black-skin-disease-dataset};
Derm12345 at \url{https://www.nature.com/articles/s41597-024-04104-3}; 
SkinCon at \url{https://skincon-dataset.github.io/}; 
Dermnet at \url{https://dermnetnz.org/};
Derm7pt at \url{https://github.com/jeremykawahara/derm7pt}; 
MSLD v2.0 at \url{https://www.kaggle.com/datasets/joydippaul/mpox-skin-lesion-dataset-version-20-msld-v20}; 
Fitz17k at \url{https://github.com/mattgroh/fitzpatrick17k}; 
DDI at \url{https://ddi-dataset.github.io/}; 
and the PAD-UFES-20 at \url{https://www.sciencedirect.com/science/article/pii/S235234092031115X}. 
A portion of the data, specifically the clinical samples used for internal validation, is not publicly available due to privacy restrictions and ethical considerations regarding patient confidentiality. Derived data supporting the findings of this study are available from the corresponding author upon reasonable request, subject to appropriate data usage agreements.

\section{Code availability}
To facilitate academic exchange while strictly adhering to data privacy and security regulations, the inference code for SkinGPT-R1 is publicly accessible at: \url{https://huggingface.co/yuhos16/SkinGPT-R1}. The full model weights are currently not publicly released to protect sensitive patient information contained within the training corpus. However, for non-commercial research purposes, access to the trained model parameters may be granted upon reasonable request to the corresponding author, subject to ethical approval and data usage agreements. We have provided comprehensive details regarding our methodology, model architecture, and training procedures in the main text and Supplementary Information to ensure the reproducibility of our experiments. Key foundational libraries utilized in this work, including PyTorch and Hugging Face Transformers, are open-source and publicly available.\\

\section{Acknowledgements}

Y.S., Z.C., Y.H., Zijian W., Y.Y., Ziwen W., X.Z., and J.Z. were supported by The Chinese University of Hong Kong, Shenzhen (CUHK-Shenzhen), under Award No UDF01004172.

\section{Credit author statement}
Y.S., Z.C., J.Z. contributed to the concept of the study and designed the research. Y.S., Z.C., S.Y., Y.H., Y.Y., Ziwen W., X.Z. collected and processed the data. Y.S., Y.H. conducted the study. Y.X., S.Z., L.S., T.L., W.L. analyzed the data. Y.S., Y.H., Zijian W., Y.Z., J.Z. co-wrote the manuscript. Y.S., S.Y., Y.Z., J.Q., Z.G., J.Z. critically revised the manuscript. Y.S., Z.C., S.Y., Y.Y., J.Z. performed the technical review. All authors discussed the results and provided comments regarding the manuscript.

\section{Competing Interests}
The authors have declared no competing interests.\\

\clearpage
{
\bibliographystyle{IEEEtran}
\bibliography{reg}
}
\clearpage

\end{document}